\documentclass[conference]{IEEEtran}

\usepackage{cite}
\usepackage{amsmath,amssymb,amsfonts}
\usepackage{graphicx}
\usepackage{textcomp}
\usepackage{xcolor}
\usepackage{booktabs} 
\usepackage{url}

\usepackage[caption=false,font=footnotesize]{subfig}

\begin{document}

\title{Trustworthy Longitudinal Brain MRI Completion: A Deformation-Based Approach with KAN-Enhanced Diffusion Model}


\author{
    \IEEEauthorblockN{
        Tianli Tao\textsuperscript{1,2,5,*}, 
        Ziyang Wang\textsuperscript{3,*}, 
        Delong Yang\textsuperscript{4}, 
        Han Zhang\textsuperscript{2,\dag}, 
        Le Zhang\textsuperscript{5,\dag}
    }
    \IEEEauthorblockA{
        \textsuperscript{1}School of Biomedical Engineering and Imaging Science, King's College London, London, UK\\
        \textsuperscript{2}School of Biomedical Engineering, ShanghaiTech University, Shanghai, China\\
        \textsuperscript{3}School of Computer Science and Digital Technologies, Aston University, Birmingham, UK\\
        \textsuperscript{4}Kunming Medical University, Kunming, China\\
        \textsuperscript{5}Digital Healthcare and Medical Imaging Research Group, School of Engineering,\\ 
        College of Engineering and Physical Sciences, University of Birmingham, Birmingham, UK
    }
}

\maketitle
\begingroup
\renewcommand\thefootnote{} 
\footnotetext{* Tianli Tao and Ziyang Wang contributed equally to this work. \dag Co-corresponding author: Han Zhang (zhanghan2@shanghaitech.edu.cn)and Le Zhang(email: l.zhang.16@bham.ac.uk).}
\endgroup
\begin{abstract}
Longitudinal brain MRI is essential for lifespan study, yet high attrition rates often lead to missing data, complicating analysis. Deep generative models have been explored, but most rely solely on image intensity, leading to two key limitations: 1) the fidelity or trustworthiness of the generated brain images are limited, making downstream studies questionable; 2) the usage flexibility is restricted due to fixed guidance rooted in the model structure, restricting full ability to versatile application scenarios. To address these challenges, we introduce DF-DiffCom, a Kolmogorov-Arnold Networks (KAN)-enhanced diffusion model that smartly leverages \textit{deformation fields} for trustworthy longitudinal brain image completion. Trained on OASIS-3, DF-DiffCom outperforms state-of-the-art methods, improving PSNR by 5.6\% and SSIM by 0.12. More importantly, its \textit{modality-agnostic} nature allows smooth extension to varied MRI modalities, even to attribute maps such as brain tissue segmentation results.
\end{abstract}



\section{Introduction}
Longitudinal Magnetic Resonance Imaging (MRI) is indispensable for tracking the brain's structural changes over the human lifespan, from rapid development in youth to gradual atrophy in aging \cite{bethlehem2022brain}. Such studies are crucial for understanding neurodegenerative diseases like Alzheimer's, where subtle changes such as hippocampal shrinkage can predate clinical symptoms \cite{pegueroles2017longitudinal}. However, longitudinal studies are plagued by participant attrition, leading to incomplete datasets that hinder robust analysis. While simple methods like linear interpolation fail to capture the brain's non-linear changes, deep generative models, such as GANs and Denoising Diffusion Probabilistic Models (DDPMs) \cite{goodfellow2020generative,ho2020denoising}, have emerged as powerful alternatives \cite{guo2024cas,zhu2024loci}.

Despite their promise, existing generative approaches for longitudinal completion suffer from two fundamental limitations. First, they are predominantly \textbf{intensity-based}, directly synthesizing pixel or voxel values. This task is inherently ill-posed, as the model must simultaneously preserve the subject's static anatomical identity and impose realistic temporal changes. This often compromises the anatomical \textbf{fidelity} of the generated images, introducing subtle artifacts or inconsistencies that can undermine their trustworthiness for sensitive downstream clinical analyzes. Second, they exhibit poor \textbf{flexibility}. Most models are designed with a fixed input structure (e.g., requiring exactly one or two prior scans), which is a poor fit for real-world clinical datasets where the number of available scans per subject is irregular.

To address these challenges, we propose a paradigm shift: instead of generating image intensities, we generate the \textit{deformation field} that maps a source image to a target time point. This approach, inspired by its success in registration tasks \cite{kim2022diffusemorph,zheng2024deformation}, offers a more principled solution. By warping a high-quality source image, we inherently preserve its detailed anatomical information, thus ensuring high fidelity. This also makes the process intrinsically \textbf{modality-agnostic}, as a valid deformation can be applied to any co-registered modality (e.g., T1w, T2w, or segmentation maps) without retraining. The core challenge then becomes learning to generate these complex, high-dimensional deformation fields. We argue that this is a function approximation problem, for which Kolmogorov-Arnold Networks (KAN) \cite{liu2024kan}, designed to excel at learning complex functions, are a more suitable backbone than generic U-Nets. Our contributions are threefold:
\begin{enumerate}
    \item We propose \textbf{DF-DiffCom}, a novel framework that generates high-fidelity deformation fields using a KAN-enhanced diffusion model, offering a more trustworthy approach to longitudinal image completion.
    \item We design a \textbf{Flexible Temporal Information Enhancement (F-TIE)} module that effectively conditions the model on a variable number of available scans, enhancing its real-world applicability.
    \item We conduct extensive experiments on the OASIS-3 dataset \cite{lamontagne2019oasis}, demonstrating that our method achieves state-of-the-art image quality.
\end{enumerate}

\section{Methods}
\begin{figure*}[!t]
    \centering
    \includegraphics[width=\textwidth]{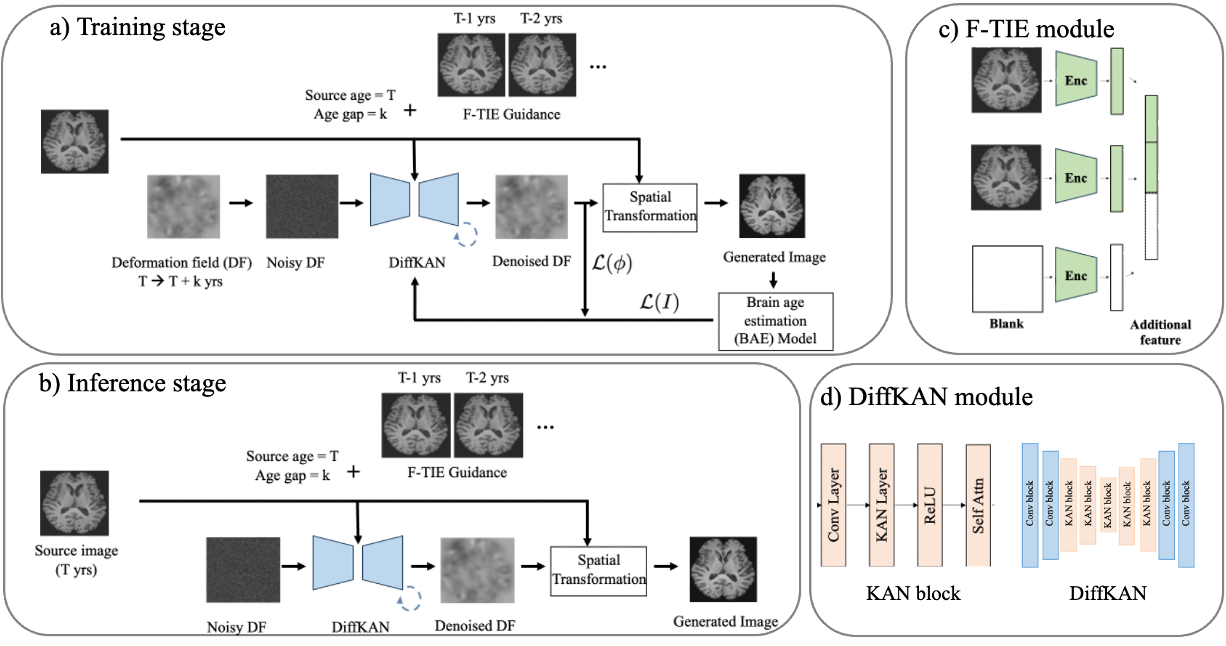}
    \caption{The schematic pipeline of our proposed DF-DiffCom. (a) During training, we use a source image, target age, and optional additional images to guide a KAN-enhanced diffusion model (DiffKAN) to predict a deformation field. A dual loss is computed on both the deformation field and the warped image. (b) During inference, the trained model generates a deformation field, which is applied to the source image via a Spatial Transformer Network (STN) to produce the completed image. (c) The Flexible Temporal Information Enhancement (F-TIE) module encodes a variable number of additional images into a fixed-length guidance vector. (d) The DiffKAN backbone integrates KAN blocks for superior non-linear feature extraction.}
    \label{fig:1}
\end{figure*}

Our framework, DF-DiffCom (Fig. \ref{fig:1}), reframes longitudinal completion. Instead of directly generating a target image $I_{target}$, we learn to generate a deformation field $\phi$ that warps a source image $I_{source}$ to the target time point: $I_{generated} = I_{source} \circ \phi$. The core of our method is a conditional diffusion model that generates $\phi$ based on the source image, target age, and other available scans.

\subsection{Deformation Field Generation via Conditional Diffusion}
We model the generation of the deformation field $\phi$ using a DDPM \cite{ho2020denoising}. The forward process gradually adds Gaussian noise to the ground-truth deformation field $\phi_0$ over $T$ timesteps. The reverse process is performed by a neural network, $\epsilon_\theta$, which is trained to predict the added noise $\epsilon$ at each timestep $t$ from the noisy field $\phi_t$. This process is conditioned on the source image $c_1=I_{source}$, the target age $t_{age}$, and an embedding from other available scans $c_2$. The model is optimized using a simplified objective:
\begin{equation}
    L_{simple} = \mathbb{E}_{t, \phi_0, \epsilon} \left[ || \epsilon - \epsilon_\theta(\phi_t, t, c_1, t_{age}, c_2) ||^2 \right],
    \label{eq:diffusion_loss}
\end{equation}
where $\epsilon \sim \mathcal{N}(0, \mathbf{I})$ is the sampled noise, and $t$ is uniformly sampled from $\{1, ..., T\}$.

\subsection{DiffKAN: KAN-enhanced U-Net Backbone}
Standard U-Nets in diffusion models use generic convolutional blocks. We hypothesize that modeling complex, non-linear deformations can be improved by a network architecture better suited for function approximation. We introduce \textbf{DiffKAN}, which replaces standard convolutional blocks in the U-Net bottleneck and expansive path with KAN-based blocks (Fig. \ref{fig:1}d). Each KAN block consists of a convolutional layer followed by a KAN layer \cite{liu2024kan}, which uses learnable activation functions on network edges, offering superior performance in fitting complex functions compared to MLPs with fixed activations. This design allows the network to more efficiently capture the intricate spatial relationships governing brain deformation.

\subsection{Flexible Temporal Information Enhancement (F-TIE)}
To handle a variable number of available scans for guidance, we introduce the F-TIE module (Fig. \ref{fig:1}c). Given a set of $k$ available auxiliary images $\{Img_1, ..., Img_k\}$ (where $0 \le k \le N$, with $N$ being the maximum supported number, set to $N=3$ in our experiments), each image is passed through a simple CNN encoder to produce a feature vector $v_i = Enc(Img_i)$. These vectors are concatenated and then projected by a linear layer to a fixed-length guidance vector $c_2$, which is fed into the diffusion model via cross-attention. If $k < N$, the remaining slots are zero-padded before concatenation.
\begin{equation}
    c_{2} = \text{Linear}(\text{Concat}[v_1, \dots, v_k, \mathbf{0}, \dots, \mathbf{0}]).
    \label{F-TIE}
\end{equation}
This simple yet effective strategy allows the model to leverage all available subject-specific temporal information, improving the consistency of generated trajectories.

\subsection{Dual Loss Function}
In addition to the standard diffusion loss $L_{simple}$ (Eq. \ref{eq:diffusion_loss}), we introduce two auxiliary losses to enforce domain-specific constraints. The total loss is $L_{total} = \lambda_1 L_{simple} + \lambda_2 L_{DF} + \lambda_3 L_{BAE}$.

\textbf{1) Deformation Field Loss ($L_{DF}$):} To ensure the generated deformation field $\phi_{pred}$ is both accurate and physically plausible, we apply a loss directly on the field, comparing it to the ground truth field $\phi_{gt}$. This loss combines a Normalized Cross-Correlation (NCC) term for structural alignment and a smoothness term $L_{Grad}$ that penalizes sharp discontinuities in the deformation:
\begin{equation}
    L_{DF} = 1 - \text{NCC}(\phi_{pred}, \phi_{gt}) + \gamma ||\nabla \phi_{pred}||^2.
    \label{eq:df_loss}
\end{equation}
Here, $\nabla$ is the spatial gradient operator, ensuring the deformation is smooth.

\textbf{2) Brain Age Estimation (BAE) Loss ($L_{BAE}$):} To ensure the generated image $I_{generated} = I_{source} \circ \phi_{pred}$ is biologically plausible for the target age, we employ a pre-trained BAE model as a guiding critic. The BAE model, a simple ResNet-based regressor, is trained to predict age from brain MRIs. Critically, to provide a stable signal during the noisy reverse diffusion process, the BAE model is pre-trained on images augmented with varying levels of Gaussian noise. The loss is the Mean Absolute Error (MAE) between the predicted age and the target age:
\begin{equation}
    L_{BAE} = |\text{BAE}(I_{generated}) - t_{age}|.
    \label{eq:bae_loss}
\end{equation}

\section{Experiments and Results}

\textbf{Dataset and Preprocessing:} We used 2,535 T1w MRIs from 634 subjects (age 42-95) from the OASIS-3 dataset \cite{lamontagne2019oasis}. Images were registered to the MNI152 template, skull-stripped, and resized to 160$\times$192$\times$224. Data was split by subject (70\% train, 20\% test, 10\% validation). To ensure a fair evaluation and prevent data leakage, the pre-trained TransMorph \cite{chen2022transmorph} (for ground truth deformation fields) and our BAE models were trained in a separate and non-overlapping subset of subjects from OASIS-3, which were excluded from all main training, validation, and testing sets.

\textbf{Implementation Details:} The model was trained for 500 epochs on an NVIDIA RTX 4090 GPU using the AdamW optimizer with a learning rate of 1e-4. The loss weights were empirically set to $\lambda_1=1.0$, $\lambda_2=0.5$, $\lambda_3=0.1$, and $\gamma=0.01$. The diffusion process used $T=1000$ timesteps with a linear noise schedule. Deformation fields were normalized to a range of [-1, 1] before training. While the KAN layers introduce a computational overhead compared to standard convolutions, the total training time for our DiffKAN model was comparable to a deeper U-Net baseline, as KAN's expressiveness allowed for a shallower network architecture.

\begin{figure*}[!t]
    \centering
    \includegraphics[width=\textwidth]{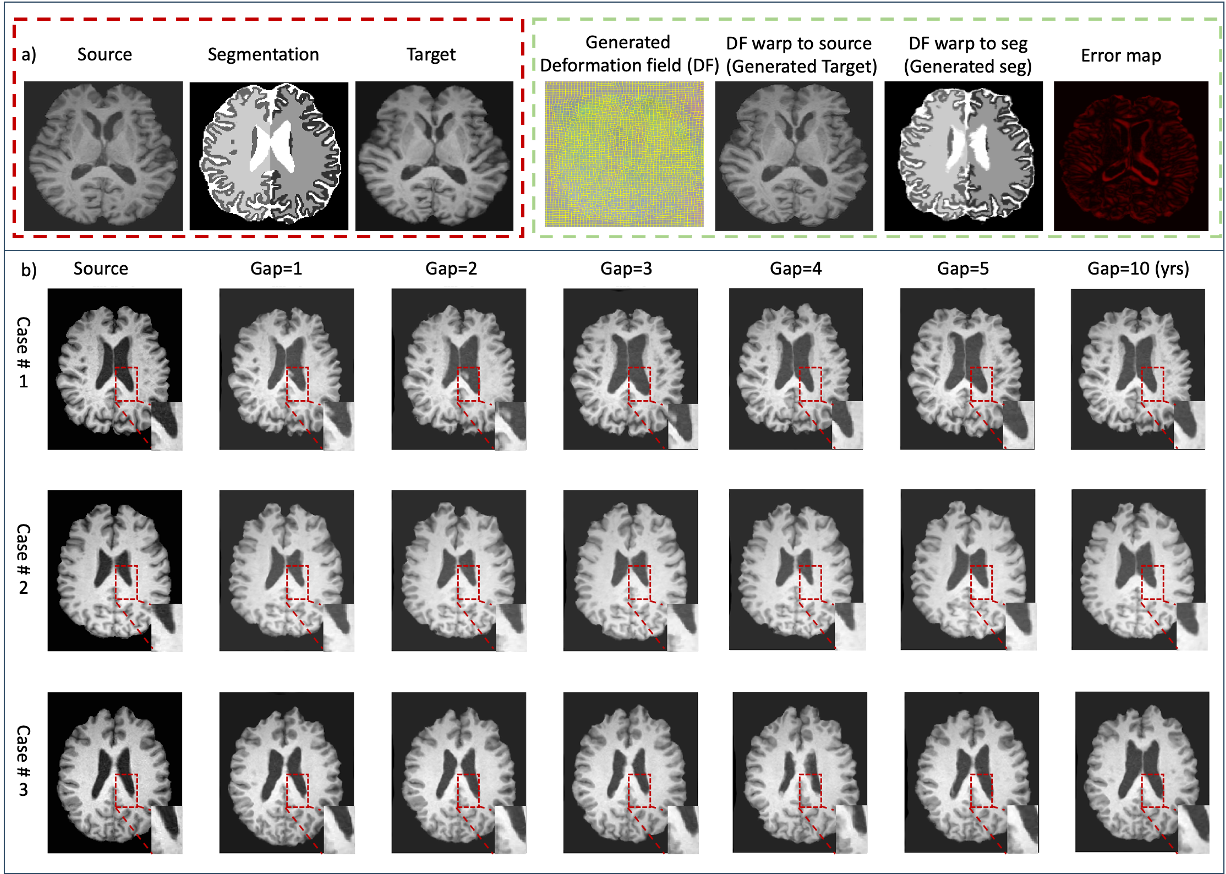}
    \caption{(a) Our method generates a deformation field (DF) to warp a source image (age 60) and its segmentation map to a target age (70). The generated images show plausible aging effects like ventricular enlargement. (b) Examples of longitudinal generation with increasing age gaps, demonstrating the model's ability to capture progressive and anatomically consistent changes, such as the gradual expansion of ventricles over time.}
    \label{fig:3}
\end{figure*}

\textbf{Qualitative Results:} As shown in Fig. \ref{fig:3}, our method generates realistic aging patterns. For instance, when generating a 70-year-old brain from a 60-year-old source MRI, the model produces plausible ventricular enlargement. The intrinsic modality-agnosticism is demonstrated by applying the same deformation field to both a T1w image and its segmentation map, yielding a consistent pair of future data points. Our model also captures complex, non-linear aging trajectories observed in real data. The apparent ventricular shrinkage in some 2D slices (as seen in some examples) can reflect non-monotonic volume changes over time or be an artifact of slice selection through a 3D structure that is globally, but not uniformly, expanding.

\textbf{Quantitative Results:} We compared DF-DiffCom with state-of-the-art methods. As shown in Table \ref{Table:1}, our method achieved the highest mean PSNR and SSIM, with lower standard deviations in SSIM, indicating superior image quality and structural fidelity.

\begin{table}[h] 
\centering
\caption{Quantitative comparison with baseline methods. Values are mean ± std. Our method shows superior performance.}
\label{Table:1}
\setlength{\tabcolsep}{4mm} 
\begin{tabular}{lcc}
\toprule
Method  & PSNR  &  SSIM     \\ 
\midrule
cGAN \cite{isola2017image}         &18.92 $\pm$ 1.58          &0.64 $\pm$ 0.08           \\ 
DiffuseMorph \cite{kim2022diffusemorph} &19.67 $\pm$ 1.51        &0.68 $\pm$ 0.07           \\
LoCI-DiffCom \cite{zhu2024loci} &20.01 $\pm$ 1.49         &0.69 $\pm$ 0.06          \\ 
TADM \cite{yoon2023sadm}           &20.51 $\pm$ 1.43         &0.72 $\pm$ 0.05           \\ 
\textbf{Ours}   &\textbf{25.52 $\pm$ 1.21}       &\textbf{0.84 $\pm$ 0.03}          \\
\bottomrule
\end{tabular}
\vspace{-3mm} 
\end{table}

\textbf{Ablation Study:} We evaluated the contribution of our two main components: the DiffKAN backbone and the F-TIE module. The results, presented in Table \ref{Table:2}, clearly demonstrate their importance. Removing both components leads to the worst performance. Individually adding either DiffKAN or F-TIE provides a notable boost in both PSNR and SSIM. The full model combining both components achieves the best results, confirming their synergistic effect and validating our design choices.

\begin{table}[h] 
\centering
\caption{Ablation study of key model components. The full model demonstrates the effectiveness of both DiffKAN and F-TIE.}
\label{Table:2}
\setlength{\tabcolsep}{3mm} 
\begin{tabular}{cccc}
\toprule
DiffKAN & Flexible guidance (F-TIE) & PSNR  &  SSIM     \\ 
\midrule
$\times$ & $\times$         &23.73±1.18             &0.794±0.04                   \\ 
$\checkmark$ & $\times$       &24.01±1.15           &0.796±0.03                  \\ 
$\times$   & $\checkmark$      &24.07±1.17          &0.798±0.03                   \\
$\checkmark$ & $\checkmark$  &\textbf{25.52±1.21}    &\textbf{0.845±0.03}          \\
\bottomrule
\end{tabular}
\vspace{-4mm} 
\end{table}

\section{Conclusion}
We introduced DF-DiffCom, a novel framework for trustworthy longitudinal brain MRI completion. By generating deformation fields with a KAN-enhanced diffusion model, our approach inherently preserves anatomical fidelity, leading to state-of-the-art image quality. The flexible guidance mechanism and intrinsic modality-agnostic nature enhance its practical applicability. While the reliance on supervised deformation fields is a current limitation, our method not only generates visually plausible images but also provides tangible benefits for downstream clinical tasks, demonstrating its potential to help build more complete and useful longitudinal datasets for neuroscience research. Future work will explore unsupervised or weakly-supervised methods for learning deformations.

\bibliographystyle{IEEEtran}
\bibliography{reference}

\end{document}